\documentclass[11pt]{article}
\usepackage[final]{acl}
\usepackage{times}
\usepackage{latexsym}
\usepackage[T1]{fontenc}
\usepackage[utf8]{inputenc}
\usepackage{microtype}
\usepackage{inconsolata}
\usepackage{graphicx}
\usepackage{xcolor}
\usepackage{array} 
\usepackage{booktabs}
\usepackage{fontawesome5}
\usepackage{amsmath}

\newcommand{\github}[1]{%
   \href{#1}{\faGithubSquare}%
}

\title{Task Arithmetic with Support Languages for Low-Resource ASR}

\author{
 \textbf{Emma Rafkin\textsuperscript{1,3\thanks{This research was completed through Georgetown and was not funded by APL. }}} \And
 \textbf{Dan DeGenaro\textsuperscript{2}} \And
 \textbf{Xiulin Yang\textsuperscript{1}} \AND \\
 \textsuperscript{1}Department of Linguistics, Georgetown University\\
 \textsuperscript{2}Department of Computer Science, Georgetown University\\
 \textsuperscript{3}Applied Physics Laboratory, Johns Hopkins University
\\
    \texttt{\{epr41, drd92, xy236\}@georgetown.edu}}

\begin{document}

\maketitle

\begin{abstract}

The development of resource-constrained approaches to automatic speech recognition (ASR) is of great interest due to its broad applicability to many low-resource languages for which there is scant usable data. Existing approaches to many low-resource natural language processing tasks leverage additional data from higher-resource languages that are closely related to a target low-resource language. One increasingly popular approach uses \textit{task arithmetic} to combine models trained on different tasks to create a model for a task where there is little to no training data. In this paper, we consider training on a particular language to be a task, and we generate \textit{task vectors} by fine-tuning variants of the Whisper ASR system. For pairs of high- and low-resource languages, we merge task vectors via a linear combination which is optimized on the downstream word error rate on the low-resource target language's validation set. Across 23 low-resource target languages for which we evaluate this technique, we find consistent word error rate improvements of up to 10\% compared to a baseline without our approach.\footnote{Code for this project can be found at \url{https://github.com/ddegenaro/mozilla-asr-challenge}.}

\end{abstract}

\section{Introduction}
\label{sec:intro}

The 2025 Mozilla Common Voice Spontaneous Speech ASR shared task focuses on developing ASR systems for 26 low-resource languages. Training data is provided for 21 languages, while the remaining five are treated as unseen. Unlike typical ASR benchmarks based on scripted speech, this task emphasizes spontaneous speech. Performance is evaluated using Word Error Rate (WER) under four tracks: overall multilingual performance (Task~1), language-specific improvement over a baseline (Task~2), baseline improvement under a 500MB model size constraint (Task~3), and generalization to unseen languages (Task~4).

In this paper, we explore the use of task arithmetic \citep{ilharco_editing_2023} to leverage higher-resource languages for low-resource ASR. We fully fine-tune the pre-trained \texttt{whisper-tiny} model \cite{radford_robust_2023} and train Low-Rank Adapters (LoRA) \cite{hu_lora_2021} for \texttt{whisper-large-v3} for each target language. To improve performance in low-resource settings, we incorporate information from genetically related higher-resource ``support languages'' \cite{nagasawa_task_2025}. For each target language, we identify closely related languages and train corresponding models on Common Voice (CV) data \cite{ardila_common_2020} using the same training setup. We then apply task arithmetic to combine target and support models (or adapters), selecting merged models based on WER on the target development sets. Our results show consistent improvements across languages, demonstrating the effectiveness of support language integration for low-resource ASR with task arithmetic.

\section{Related Work}
\label{sec:related}
\subsection{Automatic Speech Recognition Models}
Recent advances in neural architectures have led to rapid progress in ASR \citep{cui-etal-2025-recent, prabhavalkar2023end}. Speech foundation models such as wav2vec~2.0 \citep{baevski2020wav2vec}, WavLM \citep{chen2022wavlm}, and HuBERT \citep{hsu2021hubert} achieve strong performance across benchmarks with low error rates and efficient inference. Among these, OpenAI’s Whisper family has been particularly influential. Whisper models are encoder-decoder architectures trained on 680{,}000 hours of weakly labeled speech \citep{radford_robust_2023}, demonstrating that large-scale, noisy supervision can support effective multilingual ASR and speech-to-English translation.

Although Whisper is trained on nearly 100 languages, its training data is dominated by high-resource languages, leading to substantial performance disparities cross-linguistically. Nevertheless, the public release of Whisper’s model weights enables supervised fine-tuning for new languages when labeled data is available.

\subsection{Low-Resource ASR}
Low-resource ASR remains challenging due to the large amounts of labeled data required by neural models. Prior work has explored multilingual transfer to mitigate data scarcity, showing that leveraging related high-resource languages can improve recognition accuracy in low-resource settings \citep[e.g.,][]{khare2021low, ramesh_task_2024, su_task_2024, yu2023master}.

A complementary line of work uses task vectors \citep[TVs;][]{ilharco_editing_2023}, defined as the parameter-wise difference between fine-tuned and base models, which can be composed to transfer capabilities across tasks or domains. This ``task arithmetic'' can also be achieved via combining LoRA layers \citep{zhang_2023, chitale_task_2023, chronopoulou_language_2024, li_when_2025}. Recent studies apply task arithmetic to ASR \citep{ramesh_task_2024, su_task_2024}, demonstrating its utility in resource-constrained domains. \citet{nagasawa_task_2025} also use task arithmetic to improve low-resource language ASR models using high-resource counterparts, but their experiment was limited to four target languages which had all been briefly included in Whisper's training data. For this shared task, we expand their approach to more languages and language families, where most of the languages were not present during Whisper training.

\section{Methods}
\label{sec:methods}

\subsection{Models and Training}
We use LoRA layers to fine-tune \texttt{whisper-large-v3}, and we apply full fine-tuning to \texttt{whisper-tiny} to fit within the constraints of Task 3. Due to resource constraints, we quantize \texttt{whisper-large-v3} to 4-bits. We employ early stopping based on the WER on the validation set. The hyperparameters used for training can be seen in Table~\ref{tab:hyperparameters} in Appendix~\ref{hyper}.

\subsection{Leveraging Related Languages}

We make use of supplemental data from various languages which are genetically related to the languages being considered in this shared task. Following \citet{ilharco_editing_2023} and \citet{nagasawa_task_2025}, we train task vectors (or LoRA adapter-based equivalents) for each support language and then combine them with the target language models at evaluation time. 

Let $\boldsymbol{\theta}$ be the weights of the pretrained Whisper model and $\boldsymbol{\theta}_{S}$ be the weights of a model fine-tuned on ASR data from the support language(s) ${S}$. We define the TV $\boldsymbol{\tau}_{S}$ as:

\begin{equation}
    \boldsymbol{\tau}_{S} = \boldsymbol{\theta}_{S} - \boldsymbol{\theta}
\end{equation}

Note that $\boldsymbol{\theta}_{S}$ may be optimized jointly over several support languages taken in tandem. Let $\boldsymbol{\theta}_{T}$ be the weights of the model trained on ASR data in the target language. We then apply $\boldsymbol{\tau}_{S}$ to $\boldsymbol{\theta}_{T}$ to create the final model $\boldsymbol{\theta}_{final}$:
\begin{equation}
   \boldsymbol{\theta}_{final} = \boldsymbol{\theta}_{T} + \lambda \boldsymbol{\tau}_{S}
\end{equation}
Where $\lambda$ is a scaling hyperparameter.

For \texttt{whisper-large-v3}, we apply a similar process to the respective LoRA layers. We scale the weights of the layers in the support adapter, and add them to the respective target language adapter's weights to produce a merged adapter. We finally apply that merged adapter to the pre-trained Whisper model.

Support languages were chosen based on genetic relatedness to the target language, shared scripts, and the availability of scripted speech data in CV. We did not attempt any metric-based selection of support languages, as data availability in CV was already a significant limitation. We fine-tune \texttt{whisper-tiny} and train LoRA adapters for \texttt{whisper-large-v3} separately for each target language and each support language grouping. Whisper’s language detection is disabled and the language token is fixed to a \emph{Whisper proxy language}: a pre-training language selected to maximize token overlap with the target language while sharing the same script. This initialization aims to start adaptation from a representation space as close as possible to the target language. Mappings between target languages, support languages, proxy languages, and scripts are provided in Table~\ref{tab:language_maps}, with language codes listed in Tables~\ref{tab:target_lang_codes} and \ref{tab:support_lang_codes}. The selected ``high-resource'' familial languages were often still relatively low-resource. In such cases, we combine multiple related languages when possible. For example, for Puno Quechua, we use 13 Quechuan languages to construct the support models.\footnote{We were unable to identify suitable familial languages for Papantla Totonac and Toba Qom, and therefore do not use any support languages for these targets. For Scots, the closest familial language is English; in this case, we likewise omit support languages and rely on the fact that Whisper’s pre-training is heavily dominated by English.}

After training all models, we tune the task-arithmetic scaling parameter $\lambda$ on the validation set for each target language. Prior work has shown that performance is sensitive to the choice of $\lambda$ \cite{li_when_2025, nagasawa_task_2025}. Instead of a grid search, we use Bayesian optimization with ten evaluations over the range $[0,1]$. 
\subsection{Data Sources and Munging}

We use standard scripted speech data from CV for all support languages and for the five unseen target languages (Adyghe, Basaa, Kabardian, Puno Quechua, and Ushojo). For the remaining 21 target languages, we use the task-provided spontaneous speech data.

We remove all samples flagged as problematic by CV annotators, as well as samples with zero-length audio or empty transcripts. Audio longer than 30 seconds is truncated to match Whisper’s maximum input length. Transcripts are left unchanged after truncation, which may have introduced minor audio–text misalignment. Finally, we apply the transcript-cleaning script provided by the task to remove formatting irregularities.

We leverage the \texttt{votes} metadata provided in CV as a proxy for transcription quality. For the spontaneous speech datasets, transcripts typically have 0–2 votes. For the scripted speech datasets (used for unseen target languages and all support languages), CV provides separate up- and down-vote counts; we compute a single vote score as the number of upvotes minus downvotes, which is generally positive. To prioritize higher-quality data, we upsample each training sample $s$ with $v$ votes by including $v+1$ copies of $s$ in the training set. This ensures that samples with no votes are retained, while higher-confidence samples are seen more frequently during training.\footnote{Due to resource constraints, we cap each language’s dataset at 100{,}000 samples, which only affects the vote-upsampled Luganda dataset.}

\section{Results}
\label{sec:results}

\begin{figure}
    \centering
    \includegraphics[width=\linewidth]{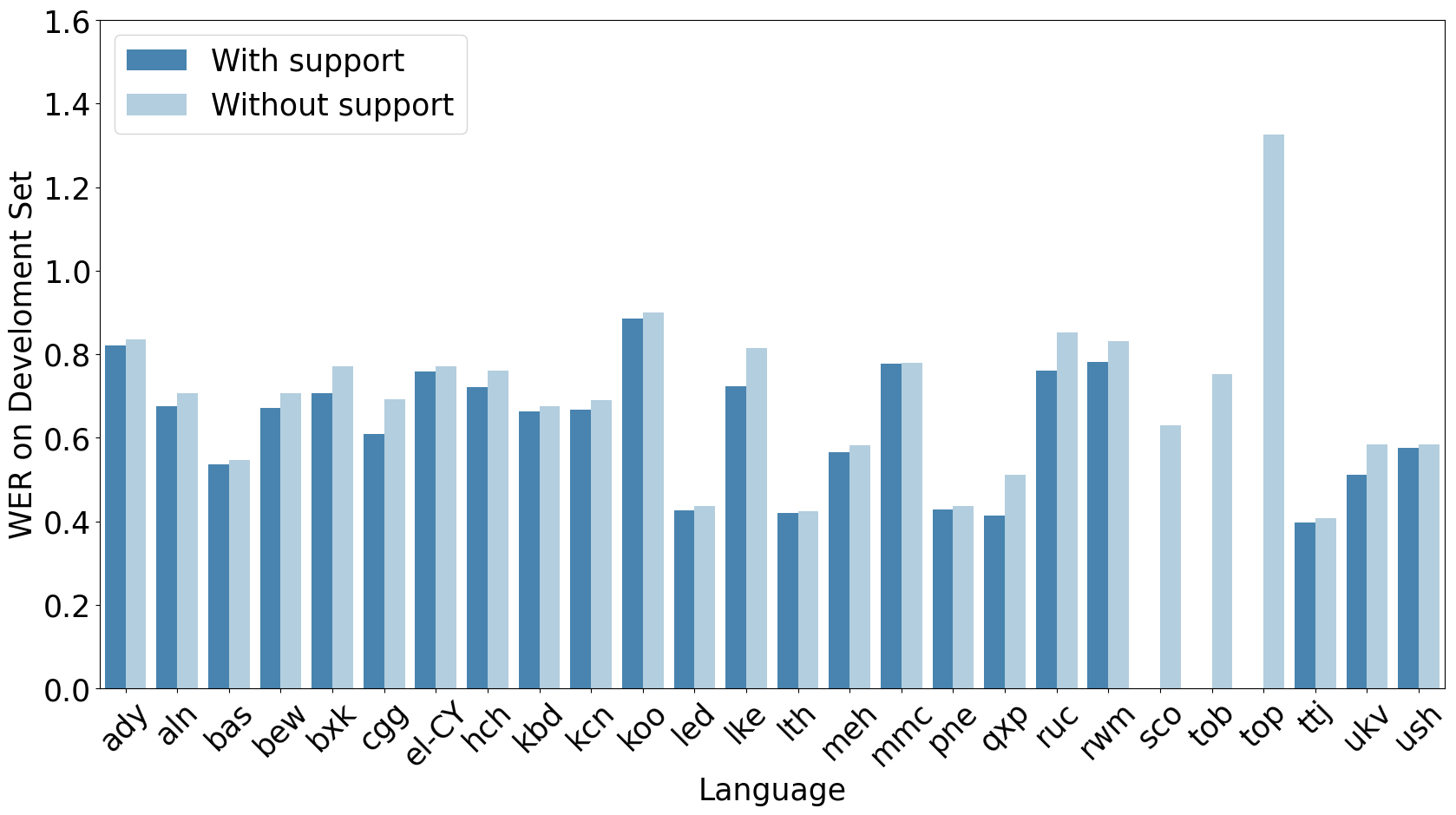}
    \caption{Incorporating the support language task vector into the \texttt{whisper-tiny} target language model improved all of the applicable target models.}
    \label{fig:whisper-tiny}
\end{figure}

Results for full fine-tuning \texttt{whisper-tiny} on the development split can be see in Figure~\ref{fig:whisper-tiny}. The development data for the unseen languages comes from the CV scripted speech corpus. We see improvements from incorporating the support language across all of the target languages, improving upon the target model by up to -9.93\% with an average WER improvement of -3.51\%. WER can be understood as the percentage of incorrect words in the transcript. Thus, depending on the performance of the original model, up to a 3-9\% decrease in WER can indicate a large improvement in an ASR model. While some language-specific improvements are small, the worst-case scenario of this approach is falling back on the baseline of only using the target language model. Experimenting with addition of a support language TV can only help. Compared to the task baseline results on the development split, our small model improves upon four of the languages: Ushojo, Toba Qom, Basaa, and Kabardian. We expect the WER to be slightly higher on the test split, specifically for the unseen language data which will be in a different domain (spontaneous speech instead of scripted speech).

\begin{figure}
    \centering
    \includegraphics[width=\linewidth]{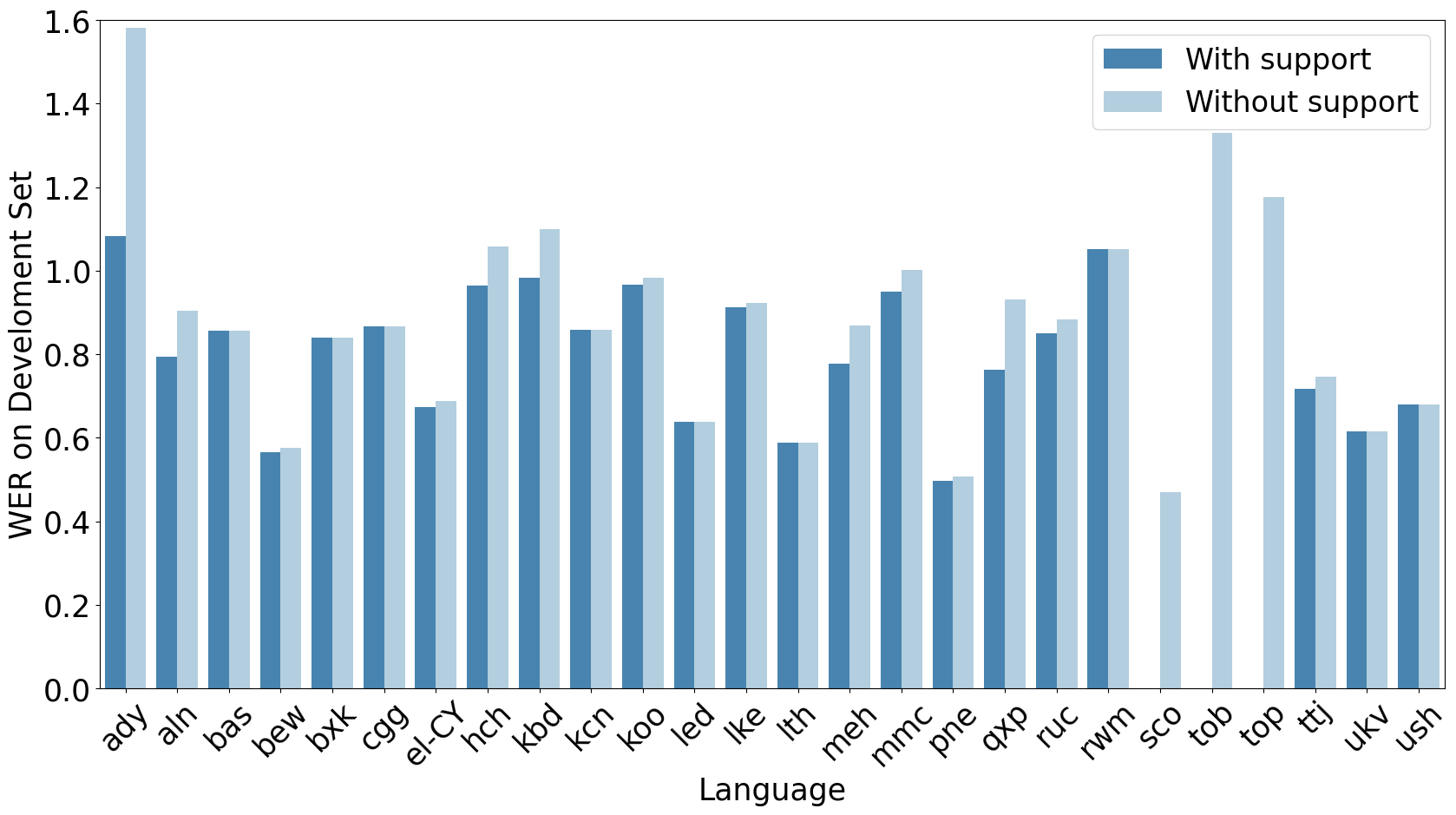}
    \caption{Incorporating the support language into the \texttt{whisper-large-v3} via linear combinations of LoRA layers led to WER improvements, though overall performance is worse than \texttt{whisper-tiny}.}
    \label{fig:whisper-large}
\end{figure}

Results from the \texttt{whisper-large-v3} model can be found in Figure~\ref{fig:whisper-large}. For the many of languages, this model performed worse than both the task development baseline and \texttt{whisper-tiny}. We only submitted \texttt{whisper-large-v3} transcripts for Scots, Papantla Totonac, and Betwi to task 1, as those were the only models that outperformed \texttt{whisper-tiny}. Additionally, there were instances in which the support language was not helpful for the target language \texttt{whisper-large-v3} model. This indicates that some of these adapters were not trained well and did not appear to capture much information from the target and support languages. Despite this, the incorporation of the support language led to average WER improvements of -5.43\%, and in the case of Adyghe this method improved upon the base model by \mbox{-49.72\%}. This indicates that task arithmetic can be successfully performed via linear combinations of LoRA layers and is not limited to TVs.

\section{Discussion}
\label{sec:discussion}

We find that using support languages consistently improved target-language performance. In contrast to findings in the original Whisper paper \cite{radford_robust_2023}, \texttt{whisper-tiny} outperformed \texttt{whisper-large-v3} on many target languages. Several factors may explain this result. Due to resource constraints, we did not perform an exhaustive hyperparameter search, and the chosen LoRA rank, quantization scheme, or other training settings may have been suboptimal for the larger model. In addition, the limited number of unfrozen parameters in \texttt{whisper-large-v3} may have restricted its capacity to adapt to the target languages. Or conversely, the small amount of available training data may have hindered generalization for the large model due to its higher overall capacity.

\begin{figure}
    \centering
    \includegraphics[width=\linewidth]{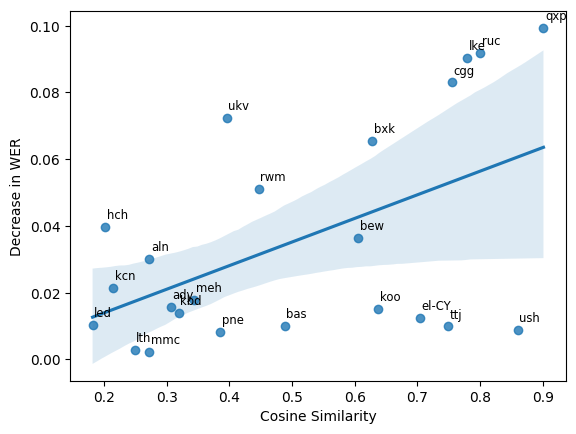}
    \caption{There is a moderate correlation between the cosine similarity of each target language and its support language and the impact of the support language TV on the target model. The shaded area is the 95\% CI.}
    \label{fig:reg_plot}
\end{figure}

The choice of support language and the scaling parameter $\lambda$ both have a substantial impact on performance. Following \citet{li_when_2025}, we hypothesize that support languages that are more closely aligned with the target language should be more beneficial. To test this, we represent each language’s training data as a vector of token counts and compute the cosine similarity between each target language and its support language. We then examine the relationship between this similarity and the relative improvement in WER from incorporating the support language ($\Delta$WER).

For \texttt{whisper-tiny}, we find a moderate positive correlation between language similarity and $\Delta$WER, with a Pearson correlation of 0.52 ($p=0.01$) and a Spearman correlation of 0.38 ($p=0.07$), as shown in Figure~\ref{fig:reg_plot}. This result suggests that support languages with greater token overlap tend to yield larger improvements. The relatively high $p$ value for the Spearman correlation means that this similarity metric does not reliably predict improvements in all cases. This calculation nevertheless appears to be a reasonable heuristic for selecting support languages, or even selecting particular corpora to be used as support data in low-resource settings. We do not observe a statistically significant relationship between language similarity and the optimal value of $\lambda$.

For \texttt{whisper-large-v3}, we do not observe a significant correlation between language similarity and either $\Delta$WER or $\lambda$. We suspect this is due to incomplete convergence of several LoRA adapters. Instead, we find a strong relationship between the baseline WER of the target-only model and the $\Delta$WER, with a Spearman correlation of 0.60 ($p<0.01$) and a Pearson correlation of 0.78 ($p<0.01$). In other words, support languages provide greater benefit when there is more room for improvement, a pattern that is particularly evident for Adyghe. This may also help explain the smaller average improvements observed for \texttt{whisper-tiny}, whose target-only models already perform relatively well.

Finally, the task baseline based on fine-tuning the Massively Multilingual Speech (MMS) model \cite{pratap2023mms} outperforms most of our systems. This suggests that MMS may be a more suitable base model than Whisper for low-resource ASR. Additionally, it is likely that some of our choices for support languages were suboptimal, and the quantitative method for determining language similarity described above might have helped us to find better support languages.

\section{Conclusion}
\label{sec:conclusion}

We describe the system for our submission to the 2025 Mozilla Common Voice Spontaneous Speech challenge. We fine-tune \texttt{whisper-tiny} and train LoRA adapters for \texttt{whisper-large-v3} on each target language and use task arithmetic with genetically related support languages to improve upon each target model. We find that this approach consistently lowers the WER on the target languages, particularly when the target language model performs poorly alone. Rather than using the largest model possible, our results show that a more targeted approach using smaller models and relevant support data can lead to greater success for low-resource ASR.


\section*{Ethical Considerations}
\label{sec:ethics}

This approach trained 74 models in total (36 for each Whisper model type), requiring a nontrivial amount of GPU hours despite the quantization and small number of parameters being tuned per model. The environmental impact of the increasing global adoption of AI requires that researchers carefully take into consideration the compute resources needed to achieve performance, and decide what quality of model is necessary for the task at hand \cite{MLSYS2022_462211f6}. 

Additionally, many of the low-resource languages described in this paper are indigenous languages. The use and impact of research on indigenous languages is oftentimes divorced from the people that speak them. We trust that the data collection methods for these languages followed ethical codes outlined by previous research regarding indigenous language data collection \cite{bird-2020-decolonising, mager-etal-2023-ethical}, and hope that the impact of the systems that we explore extends beyond general ``low-resource language research'' and into the creation of useful technology for indigenous language speakers.  

\section*{Limitations}
\label{sec:limitations}

We were unable to complete a search for the optimal training hyperparameters due to lack of a compute resources and time. In addition to training hyperparameters, the search space for $\lambda$ was relatively broad: $[0, 1]$. In the end, all of the optimal values were $<0.5$.  Even within the search space of [0, 0.5], any small change in $\lambda$ varied the performance significantly. For future work, this search space should be narrowed in order to allow for the optimal $\lambda$ to be discovered. Furthermore, as discussed in Section~\ref{sec:discussion}, the choice of base model significantly impacted the overall results of this experiment. For the best performance, a better base model should be chosen and a full hyperparameter sweep should be run for language specific training values as well as $\lambda$ values.

Another limitation lies in the constrained context window of our base model, Whisper, which is capped at 30 seconds. As a result, our WER scores were likely significantly negatively impacted on longer audio. Future work could involve adopting architectures without fixed-length constraints (e.g., Transducer-based models or CTC-based models like MMS \citep{pratap2023mms}) to better handle long-form recordings, which we anticipate would further enhance the effectiveness of our approach.

Finally, future work can improve selection criteria/prioritization of support languages and data, including comparisons of phonology, lexical borrowings, dictionary overlap, or phylogenetic tree distance.

\section*{Acknowledgments}
\label{sec:acknowledgements}

We thank Joe Garman and Christopher Cervantes for their thoughtful feedback. These experiments were conducted using Georgetown HPC. 


\bibliography{custom}

\appendix
\section{Hyperparameter Settings}
\label{hyper}
See Table~\ref{tab:hyperparameters}.

\section{Languages}
\label{sec:appendix}

Language codes can be found in Tables~\ref{tab:target_lang_codes} and \ref{tab:support_lang_codes}. Support languages for the low-resource languages are reported in Table~\ref{tab:language_maps}.

\begin{table}[!th]
    \small
    \centering
    \begin{tabular}{l|c}
    \toprule 
        \textbf{Hyperparameter} & \textbf{Value}\\
        \midrule
        LoRA rank & 32 \\
        LoRA alpha & 64\\
        LoRA dropout & 0.05\\
        LoRA target modules & \texttt{q\_proj}, \texttt{v\_proj}\\
        Batch size \texttt{whisper-tiny} & 4\\
        Batch size \texttt{whisper-large-v3} & 32 \\
        Learning rate & 5e-5\\
        Maximum epochs & 30\\
        Whisper large early stopping patience & 3 \\
        Whisper tiny early stopping patience & 5 \\
        Gradient Accumulation & 1\\
        \bottomrule
    \end{tabular}
    \caption{Training hyperparameters}
    \label{tab:hyperparameters}
\end{table} 

    \begin{table}[!tbp]
    \small
    \begin{tabular}{c|c}
        \toprule
        \textbf{Code} & \textbf{Language} \\
        \midrule 
        aln    & Gheg Albanian \\
        bew  &Betawi \\
        bxk  &  Bukusu\\
        cgg    & Chiga \\
        el-CY  &Cypriot Greek\\
        hch    &Wix\'{a}rika \\
        kcn    &Nubi \\
        koo    & Konzo \\
        led    &Lendu \\
        lke    &Kenyi \\
        lth    &Thur \\
        meh    &SW Tlaxiaco Mixtec \\
        mmc    &Michoac\'{a}n Mazahua \\
        pne   &Western Penan \\
        ruc    &Ruuli \\
        rwm    & Amba \\
        sco    &Scots \\
        tob    &Toba Qom \\
        top    &Papantla Totonac\\
        ttj    &Rutoro \\
        ukv    &Kuku\\
        \midrule
        ady & Adyghe \\
        bas & Basaa \\
        kbd & Kabardian \\
        qxp & Puno Quechua \\
        ush & Ushojo \\
        \bottomrule
        \end{tabular}
        \caption{Target language codes}

        \label{tab:target_lang_codes}
        \end{table}
        
        \begin{table}[!tbp]
        \small
        \begin{tabular}{c|c}
        \toprule
        \textbf{Code} & \textbf{Language} \\
        \midrule 
        ab    & Abkhaz \\
        cut   & Teutila Cuicatec\\
        cux   & Tepeuixila Cuicatec \\
        el    & Greek \\
        id    & Indonesian \\
        kln   & Kalenjin \\
        lg   &Luganda\\
        luo   & Dholuo\\
        mau   & Huautla Mazatec\\
        ms    & Malay \\
        mt    & Maltese\\
        ncx  & Central Puebla Nahuatl\\
        nhi   & Tetelancingo Nahuatl \\
        nlv   & Orizaba Nahuatl\\   
        qup   & Southern Pastaza Quechua \\
        qur   & Quechua Yanahuanca\\
        qus   & Quechua Santiago del Estero\\
        qux   & Quechua Yauyos\\
        quy   & Quechua Chanka\\ 
        qva   & Quechua Ambo-Pasco\\
        qvl   & Quechua Cajatambo\\
        qwa   & Quechua Corongo Ancash\\
        qws   & Quechua Sihuas Ancash\\
        qxa   & Quechua Chiquián\\
        qxt   & Quechua Pasco Santa Ana de Tusi\\
        qxu   & Quechua Arequipa-La Unión\\
        qxw   & Quechua Jauja Wanka\\
        sq   & Albanian\\
        tar   & Central Tarahumara \\
        ur    & Urdu \\
        var   & Huarijio\\
        yaq   & Yaqui \\

        \bottomrule
        \end{tabular}
        \caption{Support language codes}
        \label{tab:support_lang_codes}
        \end{table}
        
       \begin{table*}[!th]
\small
    \centering
    \begin{tabular}{c|  >{\centering\arraybackslash} p{4cm} | >{\centering\arraybackslash} p{2cm}|c|c}
    \toprule
        \textbf{Target Language} & \textbf{Language Family} & \textbf{Support Language} & \textbf{Whisper Proxy} & \textbf{Script} \\
        \midrule 
        bxk       & Bantu, Zone J & lg & Swahili & Latin \\
        cgg       & Bantu, Zone J & lg & Swahili      & Latin \\
        koo       & Bantu, Zone J & lg & Swahili     & Latin \\
        lke       & Bantu, Zone J & lg & Swahili & Latin \\
        ruc       & Bantu, Zone J & lg & Swahili      & Latin \\
        ttj       & Bantu, Zone J & lg & Swahili      & Latin \\
        rwm       & Bantu, Zone D & lg & Swahili      & Latin \\
        bas*      & Bantu, Zone A & lg & Swahili      & Latin \\
        \midrule
        kcn       & Afro-Asiatic, Semitic & mt & Maltese  & Latin\\
        \midrule
        led       & Nilo-Saharan, E. Sudanic, Central & luo, kln & Somali  & Latin \\
        lth       & Nilo-Saharan, E. Sudanic, W. Nilotic &  luo, kln  & Somali & Latin \\
        ukv       & Nilo-Saharan, E. Sudanic, E. Nilotic & luo, kln  &  Somali  & Latin\\
        \midrule
        hch       & Uto-Aztecan & ncx, nhi, nlv, yaq, tar, var  & Spanish  & Latin\\
        \midrule
        meh       & Oto-Manguean, Mixtecan & cut, cux, mau & Spanish & Latin\\
        mmc       & Oto-Manguean, Otomian &  cut, cux, mau &  Spanish  & Latin\\
        \midrule
        top       & Totonacan & $\emptyset$ & Spanish & Latin\\
        \midrule
        qxp*      & Quechuan & qu\{p,x,r,y,s\}, qx\{a,u,w,t\}, qv\{l,a\}, qw\{a,s\} & Spanish  & Latin\\
        \midrule
        tob       & Guaicuruan & $\emptyset$ & Spanish & Latin\\
        \midrule
        aln       & Indo-European, Albanian           & sq          & Albanian &  Latin \\
        el-CY     & Indo-European, Greek              & el          & Greek    & Greek \\
        sco       & Indo-European, Germanic, West     & $\emptyset$ & English  & Latin \\
        ush*      & Indo-European, Indo-Aryan, Dardic & ur          & Urdu  & Perso-Arabic \\
        \midrule
        ady*      & Northwest Caucasian & ab & Kazakh & Cyrillic \& Latin\\
        kbd*      & Northwest Caucasian & ab & Kazakh & Cyrillic \& Latin \\
        \midrule
        bew       & Austronesian, Malay & id, ms & Malay  & Latin \\
        pne       & Austronesian, Malayo-Polynesian, Kenyah & id, ms & Malay  & Latin\\
        \bottomrule
    \end{tabular}
    \caption{Low-resource target languages, their selected support counterparts and Whisper proxy language codes. * indicates a ``test-only'' or ``unseen'' language. Data sourced from each language's Wikipedia page.}
    \label{tab:language_maps}
    \end{table*}
\end{document}